\documentclass[10pt,twocolumn,letterpaper]{article}

\usepackage[pagenumbers]{cvpr} 

\usepackage[dvipsnames]{xcolor}
\usepackage{graphics} 
\usepackage{epsfig} 
\usepackage{times} 
\usepackage{amsmath} 
\usepackage{amssymb}  
\usepackage{enumitem}
\usepackage{mathtools}
\usepackage{algorithm}
\usepackage{algorithmic}
\usepackage{wrapfig}
\usepackage{colortbl}
\usepackage{listings}
\usepackage{afterpage}
\usepackage{relsize}
\usepackage{multicol}
\usepackage{multirow}
\usepackage{wasysym}
\usepackage{booktabs}
\usepackage{caption}
\usepackage{subcaption}
\usepackage{sidecap}
\usepackage{pifont}
\usepackage[nodisplayskipstretch]{setspace}
\usepackage{xcolor}
\usepackage{url}

\definecolor{myblue}{HTML}{3E3EE9}
\definecolor{myorange}{HTML}{FF9000}
\definecolor{mygreen}{HTML}{39BE28}
\definecolor{myyellow}{HTML}{D1CA00}

\newcommand{\gcmark}{\textcolor{mygreen}{\ding{51}}}

\newcommand{\rxmark}{\textcolor{red}{\ding{55}}}

\newcommand{\ourwork}{ORG\xspace}

\newcommand\mypar[1]{\par\vspace{-0.2mm}\noindent\textbf{#1}\;\;}
\definecolor{LightGrey}{rgb}{0.92,0.92,0.92}
\definecolor{Myred}{rgb}{1.00,0.12,0.36}
\definecolor{Myblue}{rgb}{0,0.60,0.87}

\definecolor{cvprblue}{rgb}{0.21,0.49,0.74}
\usepackage[pagebackref,breaklinks,colorlinks,citecolor=cvprblue]{hyperref}

\def\paperID{10423} 
\def\confName{CVPR}
\def\confYear{2024}

\title{\vspace{-7mm}Floating No More: Object-Ground Reconstruction from a Single Image}

\author{
Yunze Man$^1$,
Yichen Sheng$^{2}$,
Jianming Zhang$^{3}$,
Liang-Yan Gui$^{1}$,
Yu-Xiong Wang$^{1}$
\\[0.3em]
$^1$University of Illinois Urbana-Champaign \ \ \
$^2$Purdue University \ \ \ 
$^3$Adobe
\\[0.2em]
{\tt\small \{yunzem2,lgui,yxw\}@illinois.edu, sheng30@purdue.edu, jianmzha@adobe.com}
}

\begin{document}

\twocolumn[{%
\renewcommand\twocolumn[1][]{#1}%
\maketitle
\begin{center}
    \centering
    \captionsetup{type=figure}
    \includegraphics[trim=430 1170 360 890, clip=True, width=1\textwidth]{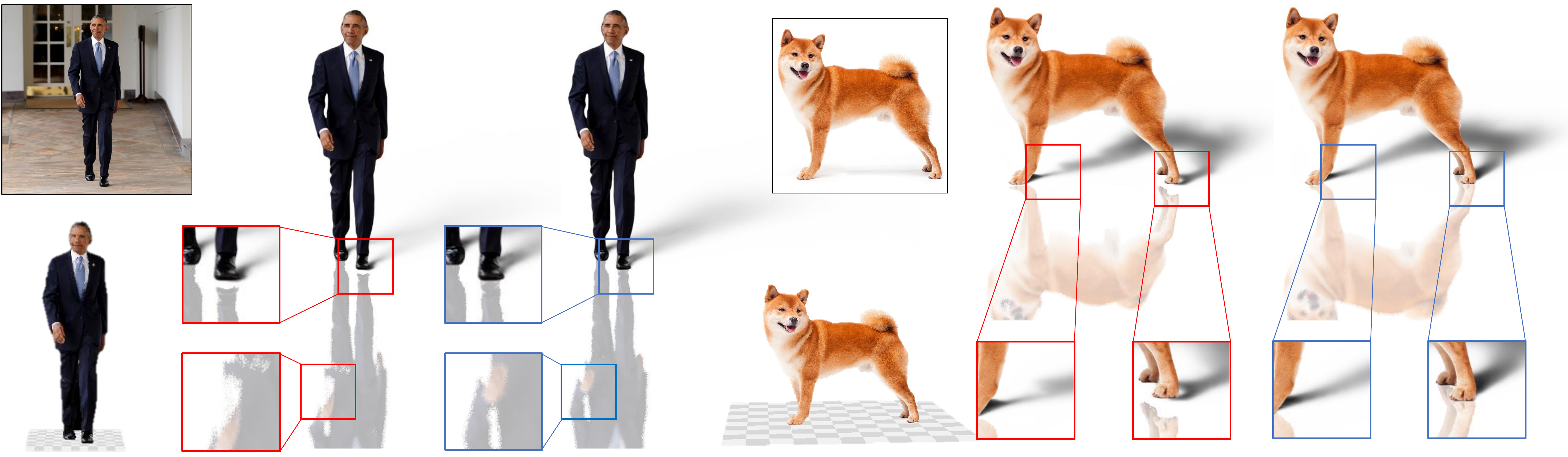}
    \vspace{-6mm}
    \captionof{figure}{Our proposed \textbf{\ourwork} (Object Reconstruction with Ground) model simultaneously reconstructs a 3D object, estimates camera parameters, and models the object-ground relationship from a monocular image. During shadow and reflection generation, the prior depth-based object geometry estimation method can result in floating issue or an unnatural shadow on the ground, as demonstrated in \textcolor{Myred}{red} boxes. Our method, on the other hand, achieves significantly more realistic editing and generation, as shown in \textcolor{Myblue}{blue} boxes. 
    }
    \label{fig:teaser}
\end{center}
}]

\begin{abstract}
\vspace{-3mm}
Recent advancements in 3D object reconstruction from single images have primarily focused on improving the accuracy of object shapes. Yet, these techniques often fail to accurately capture the inter-relation between the object, ground, and camera. As a result, the reconstructed objects often appear floating or tilted when placed on flat surfaces. This limitation significantly affects 3D-aware image editing applications like shadow rendering and object pose manipulation. 
To address this issue, we introduce \ourwork (Object Reconstruction with Ground), a novel task aimed at reconstructing 3D object geometry in conjunction with the ground surface. Our method uses two compact pixel-level representations to depict the relationship between camera, object, and ground. Experiments show that the proposed \ourwork model can effectively reconstruct object-ground geometry on unseen data, significantly enhancing the quality of shadow generation and pose manipulation compared to conventional single-image 3D reconstruction techniques.

\end{abstract}

\section{Introduction}\label{sec:intro}

The task of reconstructing an object in conjunction with a physically plausible ground, while not extensively explored, is of significant importance. This is particularly relevant in the realm of image editing applications, where it influences key aspects like controllable shadow/reflection synthesis and object view manipulation. In this work, we aim at predicting an accurate and grounded representation of objects in 3D space from a single image, specifically under unrestricted camera conditions. Recent single-view approaches have demonstrated considerable promise in tackling object reconstruction~\cite{ranftl2020midas,yin2021leres,liu2023zero,xu2023neurallift,saito2020pifuhd}. However, due to the lack of integrated object-ground modeling, objects reconstructed using these methods often appear to be ``floating'' or tilted when placed on a flat surface, which greatly hinders the realistic rendering.

\begin{figure}[!t]
    \centering
    \includegraphics[trim=620 1377 590 1234, clip=True, width=\linewidth]{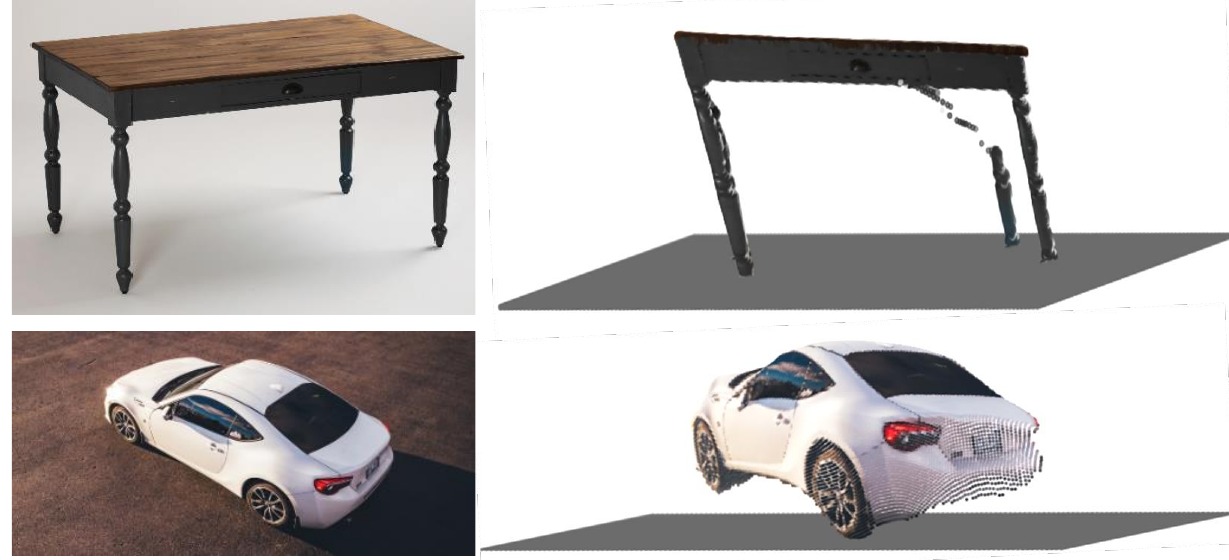}
    \vspace{-4mm}
    \caption{Without modeling object-ground correlation, existing single-view 3D estimation method~\cite{ranftl2020midas} generates 3D models floating or tilted on the ground. 
    }
    \label{fig:bad-ground}
    \vspace{-5mm}
\end{figure}

More specifically, recent works on \textit{monocular depth estimation}~\cite{chen2016single,chen2019learning,ranftl2020midas,yin2021leres} has shown great performance. They aim to recover the 3D information of an object from a single-view image by directly estimating the pixel-level depth values. Their models have been trained on large-scale datasets, and thus can generalize well on in-the-wild images. However, as pointed out by \cite{yin2021leres}, to project the depth map into 3D point clouds, additional camera parameters are needed. In some cases, off-the-shelf estimators can provide a rough estimate of these parameters, but this approach can limit the flexibility and effectiveness of object reconstruction in uncontrolled environments. Moreover, the unknown shift in the depth or disparity map will cause distortion in the 3D reconstruction (see Figure~\ref{fig:bad-ground} top row). Without an explicit modeling of the object-ground relationship, recovered 3D objects are often hard to place on a flat support plane (see Figure~\ref{fig:bad-ground} bottom row). These challenges are also present in recent category-specific 2D-to-3D methods that recover 3D shape from \textit{latent embedding space}~\cite{cheng2022sdfusion,saito2019pifu,saito2020pifuhd,wang2018pixel2mesh,wu2018learning} and zero-shot novel-view synthesis methods~\cite{liu2023one2345,liu2023zero,qian2023magic123,melaskyriazi2023realfusion,tang2023makeit3d,xu2023neurallift}, where they often just assume a simple orthographic camera model, or assume the camera parameter being given as input to avoid over-complication of the problem, which on the other hand limits their application in unconstrained scenarios.

To address these challenges, we propose \ourwork (Object Reconstruction with the Ground), a new formulation for representing objects in relation to the ground. Given a single image, our objective is to simultaneously deduce the 3D shape of the object, its positioning relative to the ground plane, and the camera parameters. We compare our method with three existing research strands: depth estimation, latent embedding reconstruction, and diffusion-based novel-view synthesis, in addition to multi-view reconstruction techniques, as detailed in~\Cref{tab:intro-comparison}. Existing single-view methods often fail to maintain the object-ground relationship and usually presuppose known camera parameters or rely on overly simplistic camera models, leading to suboptimal performance for tasks like efficient shadow generation. In stark contrast, the output of our model supports the intricate interplay between object, ground, and camera (see Figure~\ref{fig:teaser}), facilitating superior shadow generation and pose-aware geometric reconstruction. To this end, we model the object as consisting of its front (visible) and back surfaces, and predict two pixel-level height map between the object and the ground~\cite{sheng2022controllable}, along with a dense camera parameter descriptor~\cite{jin2022perspective}. Our results demonstrate that such a simplified representation of objects is not only adequate for generating 3D-realistic shadows but also yields convincing reconstruction for a wide array of commonly encountered objects.

\setlength{\tabcolsep}{3pt}
\begin{table}[!t]
\vspace{0mm}
\normalsize
\centering
\resizebox{0.97\linewidth}{!}{
\begin{tabular}{@{\hskip 1mm}r@{\hskip 2mm}c@{\hskip 4mm}c@{\hskip 4mm}c@{\hskip 4mm}c@{\hskip 4mm}c@{\hskip 2mm}}
\toprule
    &  Multi-view  &  Latent &  Depth   &  NVS  & \textbf{Ours} \\ 
\midrule
single image        & \rxmark       & \gcmark       & \gcmark        &\gcmark& \gcmark \\
category-free       & \gcmark       & \rxmark       & \gcmark        &\gcmark& \gcmark \\
camera-aware        & \gcmark       & \rxmark       & \rxmark        &\rxmark& \gcmark \\
ground-aware        & \rxmark       & \rxmark       & \rxmark        &\rxmark        & \gcmark \\
\bottomrule
\end{tabular}
}
\vspace{-2mm}
\caption{\ourwork processes multiple advantages from flexibility to generalization against multi-view reconstruction work and other single-view work including generation from latent embedding, monocular depth estimation, and novel-view synthesis (NVS) methods.}
\label{tab:intro-comparison}
\vspace{-2mm}
\end{table}

We create our training data from \textit{Objaverse}~\cite{objaverse}, rendering six images for each object with diverse focal length and camera viewpoint. We evaluate our method across two unseen datasets, including objects and humans, and show qualitative results on random unseen web images. Our proposed method outperforms existing methods in terms of accuracy, robustness, and efficiency in various scenarios. Results show that our method achieves superior performance and provides a more comprehensive and light-weight solution to the challenges of single-view object geometry estimation. In summary, our main contributions are as follows. 
\begin{itemize}[]
    \item A novel framework \ourwork, for in-the-wild single-view object-ground 3D geometry estimation. To the best of our knowledge, this is the first method to jointly model object, camera, and ground plane from single image. 
    \item  We propose a perspective field guided pixel height re-projection module to efficiently convert our estimated representations into common depth maps and point clouds.
    \item \ourwork achieves outstanding shadow generation and reconstruction performance on unseen real-world images, demonstrating great robustness and generalization ability.
\end{itemize}

\section{Related Work}\label{sec:related}

\noindent\textbf{Single-view Depth Estimation. }
There has been significant progress made in recent times in the estimation of monocular depth~\cite{chen2019learning,chen2020oasis,eigen2014depth,ranftl2020midas,wang2019web,yin2021leres}. Given metric depth supervision, some work directly trains their model to regress the depth objective~\cite{eigen2014depth,liu2015learning,yin2019enforcing,yin2021leres}. While these methods achieve great performance on various datasets, the difficulty of obtaining metric ground truth depth hinders the use of direct depth supervision. Instead, another line of work relies on ranking losses, which evaluates relative depth~\cite{chen2016single,xian2018monocular}, or scale- and shift-invariant losses~\cite{ranftl2020midas,wang2019web} for supervision. The latter methods produce particularly robust depth predictions without heavy annotation efforts, but the models are not able to reason object-ground relationship and often produce unrealistic results when using depth map for downstream image editing tasks. In light of this, a recent work~\cite{sheng2022controllable} proposes another annotation-friendly representation, pixel height, for better object shadow generation. However, this method has strict constraints on the camera viewpoint. We repurpose the representation for monocular 3D reconstruction and loosen the viewpoint by joint modeling camera with object geometry.

\vspace{1.5mm}
\mypar{Single-view 3D Geometry Reconstruction.}
Reconstructing object shapes from single-view image is a challenging but well-established problem, with one of its seminal work~\cite{roberts1963machine} optimizing for 6-DoF poses of objects with known 3D models. In the ensuing decades, learning-based methods have begun to propose category-specific networks for 3D estimation that span a wide range of objects with~\cite{cashman2012shape,kar2015category} and without direct 3D supervision~\cite{goel2020shape,kanazawa2018learning,li2020self,ye2021shelf}, and using neural implicit representations~\cite{mescheder2019occupancy,ye2022s}. With robust 3D supervision, recent methods have demonstrated the feasibility of learning 3D geometry with limited memory. Pixel2Mesh~\cite{wang2018pixel2mesh} offers a method to reconstruct the 3D shape with mesh using a single image input. Meanwhile, PIFu~\cite{saito2019pifu,saito2020pifuhd} offers an efficient implicit function to recover high-resolution surfaces of humans, including previously unseen and occluded regions. While achieving great performance, some of these works rely on learning priors specific to a certain object category, hindering its generalizabilty in the wild. Recently, advances in text-to-3D generation~\cite{poole2022dreamfusion,chen2023fantasia3d,lin2023magic3d,wang2023prolificdreamer} also inspire image-to-3D generation using diffusion prior~\cite{liu2023one2345,liu2023zero,qian2023magic123,melaskyriazi2023realfusion,tang2023makeit3d,xu2023neurallift}. Masked autoencoders are also used to object reconstruction from single image~\cite{wu2023mcc}.
In comparison, our method is the first one to model the object geometry with respect to the ground for efficient image editing and 3D reconstruction.

\vspace{1.5mm}
\mypar{Camera Parameter Estimation.}
An essential aspect of single-view monocular 3D object comprehension is to retrieve the focal length of a camera and the camera pose relative to the object and the ground plane. Classic methods leverage reference image components, including calibration grids~\cite{zhang2000flexible} or vanishing points\cite{deutscher2002automatic}, to estimate camera parameters. Recently, data-driven approaches have been proposed to use deep neural networks to infer the focal length~\cite{hold2018perceptual,workman2015deepfocal} and camera poses~\cite{lee2021ctrl,man2019groundnet,xian2019uprightnet} directly from in-the-wild images, or to use dense representation~\cite{jin2022perspective} to encode camera parameters for a more robust estimation. In contrast, our method \ourwork jointly estimates intrinsic and extrinsic camera parameters together with object geometry and ground positions, achieving a self-contained pipeline for 3D-aware image editing and reconstruction.

\begin{figure*}[!t]
    \centering
    \includegraphics[trim=790 1200 1050 1000, clip=True, width=\linewidth]{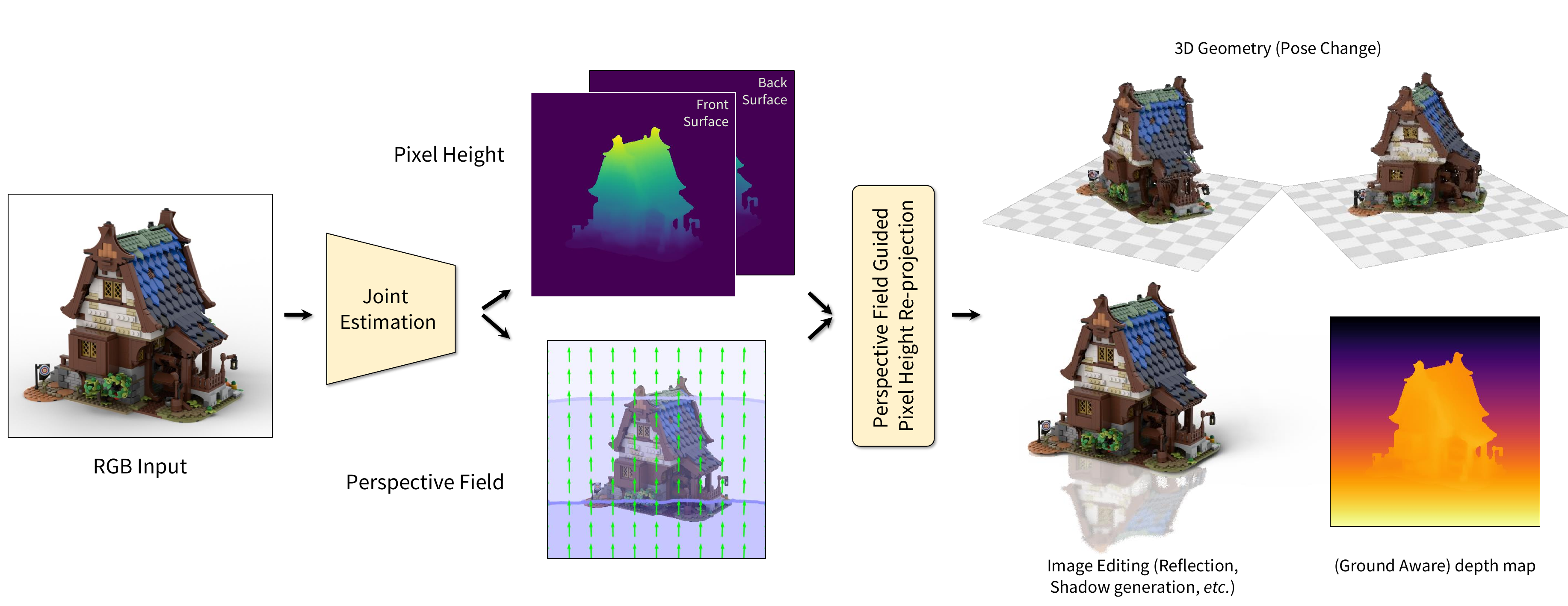}
    \vspace{-6mm}
    \caption{\textbf{\ourwork Paradigm}. Our proposed method is able to take a single-view object-centric image as input, and jointly estimate two dense representations, the pixel height and perspective field, encoding the object-ground relationship and camera parameters, respectively. A Perspective Field Guided Pixel Height Re-projection module is proposed to repurpose the two predicted dense fields into depth map estimation and point cloud generation.
    }
    \label{fig:main-pipeline}
    \vspace{-1mm}
\end{figure*}

\section{Approach}\label{sec:approach}

\ourwork considers single-view object geometry estimation by joint \textit{pixel height} and \textit{perspective field} prediction. We provide an overview of our framework in Figure~\ref{fig:main-pipeline}. Modeling object geometry and camera parameters as dense fields, we first introduce the background knowledge of the dense object-ground and dense camera representations (Sec~\ref{sec:approach:bg}). We learn a pyramid vision transformer (PVT)~\cite{wang2021pyramid,wang2022pvt} to predict the dense representation fields (Sec~\ref{sec:approach:net}), and prove that they can be repurposed for reconstruction task by proposing a perspective-guided pixel height reprojection method (Sec~\ref{sec:approach:geometry}).

\subsection{Object, Ground, and Camera Representations}\label{sec:approach:bg}

\noindent\textbf{Pixel Height Representation. } Proposed for single-image shadow generation~\cite{sheng2022controllable, sheng2023pixht}, pixel height is a dense representation defined as the pixel distance between a point on an object and its ground projection, namely its vertical projection on the ground in the image, as we can see in Figure~\ref{fig:main-pipeline}. It is a pixel-level scalar which measures the distance between object and its supporting plane in the image coordinate (number of pixels, rather than meters). Pixel height possesses many advantages over the depth representation in modeling the object geometry. First, it is disentangled from the camera model, and thus can be directly inferred from the images context without additional camera information. Moreover, it models the object and ground relationship, which is pivotal in generating realistic 3D models for real-world image applications, as objects almost always have a canonical position on the ground plane.

While photo-realistic shadows can be generated from pixel height map with projective geometry, we see more potential in this new representation. Constraining the object location with respect to a 2D plane, the pixel height representation plays a critical role in reconstructing 3D shape of objects on top of the ground. Moreover, strict requirements are enforced on camera viewpoints for pixel height~\cite{sheng2022controllable}, and only the front surface of a object is considered. Therefore, we propose to loosen this condition by modeling both front and back surfaces of the object, and jointly predicting camera intrinsic and pose relative to the ground. In the end, the field-of-view (FoV) is used to lift pixel distances into metric distance, and camera viewpoint helps align the object into the canonical pose relative to the ground.

\vspace{1.5mm}\mypar{Perspective Field Representation.} As shown in Figure~\ref{fig:main-pipeline}, the perspective field representation of a given image is composed of two dense fields, a latitude field represented by blue contour lines, and a up-vector field represented by green arrows~\cite{jin2022perspective}. Specifically, assuming a camera-centered spherical coordinate system where the zenith direction is opposite to gravity. The camera model $\mathcal{K}$ projects a 3D position $X\in \mathbb{R}^3$ in the spherical coordinate into the image frame $x\in \mathbb{R}^2$. For each pixel location $x$, the up-vector is defined as the projection of the tangential direction of $X$ along the meridian towards the north pole, the latitude is defined as the angle between the vector pointing from the camera to $X$ and the ground plane. In other words, the latitude field and the up-vector field encode the elevation angle and the roll angle of the points on the object, respectively. Both perspective fields and pixel height map are invariant or equivariant to image editing operations like cropping, rotation and translation. As a result, they are highly suitable for neural network models designed for dense prediction tasks.

\subsection{Dense Field Estimation}\label{sec:approach:net}

We present a neural network model to estimate the two dense fields from a single image. The per-pixel structure and translation-invariant nature of the pixel height and perspective field representations make them highly suitable for neural network prediction. Following~\cite{ranftl2020midas,yin2021leres}, we formulate the dense field estimation task as a regression problem. Specifically, for each image pixel of the pixel height field, assuming a ray starting from a camera pointing towards the pixel travels through the object, there will be an entry point on the front surface of the object $\mathbf{p}_\mathrm{f}$ and an exit point on the back surface of the object $\mathbf{p}_\mathrm{b}$. When the ray passes the surfaces of object multiple times, we only consider the first entry and last exit. The model is then asked to predict the pixel height for both $\mathbf{p}_\mathrm{f}$ and $\mathbf{p}_\mathrm{b}$. Moreover, we normalize it with the height of the input image. For latitude field, we normalize the original $[-\pi/2, \pi/2]$ range into $[0, 1]$. And for the up-vector field, each angle $\theta$ can range from $0$ to $2\pi$, so direct normalization and regression pose ambiguity for the model since $0$ and $2\pi$ represent the same angle. Hence, we represent each angle $\theta$ with a tuple $(\sin{\theta}, \cos{\theta})$, and train the model to regress to a two-channel vector map. All regression tasks are trained with loss $\ell_2$.

\vspace{1.5mm}\mypar{Model Architecture and Training Details.} We use the architecture of PVTv2-b3~\cite{wang2022pvt} as our backbone to extract joint feature map. We use SegFormer~\cite{xie2021segformer} with the Mix Transformer-B3 as our decoder. Residual connection is added before the decoder to include lower-level context from the 2-layer CNN block. We find that transformer-based encoder is suitable for our task as it effectively maintains global consistency in the two dense representation fields. We further make modifications to the decoder head, enabling it to produce a regression value for the pixel height map, up field map, and latitude field map. We use PVTv2-b3 pretrained on COCO dataset~\cite{lin2014microsoft} as the backbone of our architecture. The model is trained with AdamW~\cite{loshchilov2017decoupled} optimizer with learning rate $0.0005$ and weight decay 1$e$-2 for $60$K steps with batch size 8 on a 4-A100 machine. We schedule the multi-step training stages at step $30$K, $40$K, and $50$K, with learning rate decreases by 10 time at each stage. We resize the image to $(512, 512)$ before using horizontal flipping, random cropping, and color jittering augmentation during training.

\subsection{Perspective-Guided Pixel Height Reprojection}\label{sec:approach:geometry}

After predicting two dense representations, we prove that they encode sufficient information to be efficiently converted into depth maps and point clouds for downstream tasks and for fair comparison with existing methods. First, since the perspective field can be generated from camera parameters, we discretize the continuous parameter range and use a grid search optimization strategy to estimate camera field-of-view $\alpha$ and extrinsic rotation matrix $\mathbf{R}$ as row and pitch angles. Afterwards, the camera focal length is calculated as $f=\frac{H}{2\tan{\alpha/2}}$, where $H$ is the height of the input image. Then the intrinsic matrix $\mathbf{K}$ is also estimated as:
{
\begin{align}\label{eq:intrinsic}
    K = \begin{bmatrix}
f & 0 & c_x \\
0 & f & c_y \\
0 & 0 & 1
\end{bmatrix},
\end{align}}%
where $(c_x, c_y)$ is the principle point of the image and is usually estimated to be the center of the image.

\begin{figure}[!t]
    \centering
    \includegraphics[trim=200 145 250 100, clip=True, width=1.\linewidth]{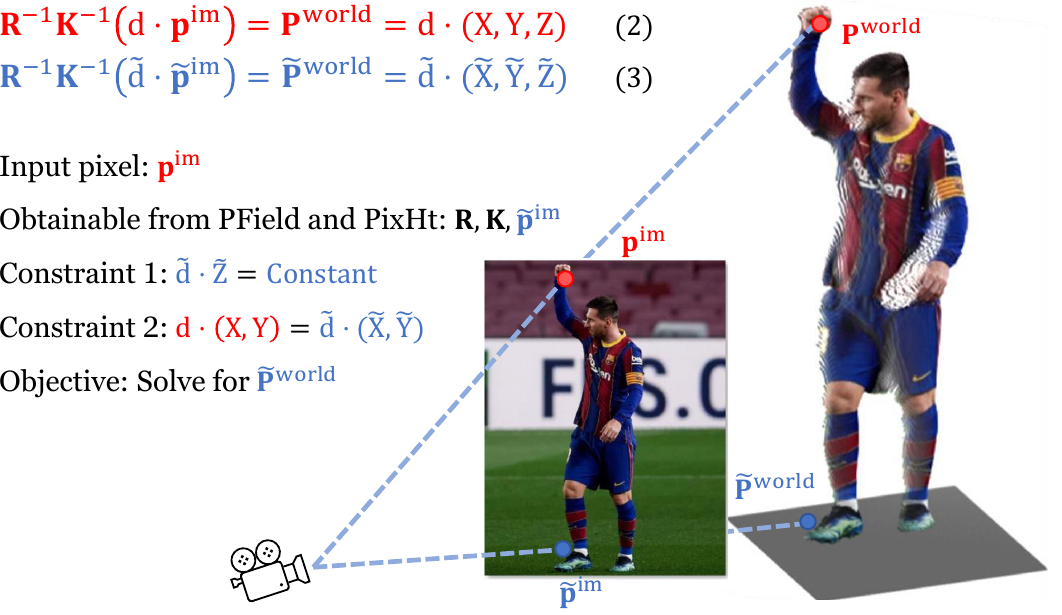}
    \vspace{-6mm}
    \caption{Perspective-Guided Pixel Height Reprojection. PField and PixHt are perspective field and pixel height, respectively.}
    \label{fig:equation}
    \vspace{-5mm}
\end{figure}

\vspace{1.5mm}\mypar{Derivation.} An illustration is provided in~\Cref{fig:equation}. Given one pixel $\mathbf{p}^{\mathrm{im}} = (x, y) \in \mathbb{R}^2$, we known its vertical projection point $\mathbf{\tilde p}^{\mathrm{im}} = (\tilde x, \tilde y) \in \mathbb{R}^2$ on the ground in the image frame, given by the estimated pixel height map. Recall that intrinsic and extrinsic matrices can be used to project a 3D point $\mathbf{P}_i$ in the world coordinate into an image pixel $\mathbf{p}_i$. More specifically, given intrinsic matrix $\mathbf{K}$ and extrinsic rotation matrix $\mathbf{R}$, we have the following equations describing the correspondence between a pixel $\mathbf{p}^{\mathrm{im}}$ on the object and its corresponding 3D points $\mathbf{P}^{\mathrm{world}}$ in the world coordinate:

{\small
  \setlength{\abovedisplayskip}{-6pt}
  \setlength{\belowdisplayskip}{6pt}
  \setlength{\abovedisplayshortskip}{0pt}
  \setlength{\belowdisplayshortskip}{3pt}
 \begin{align}
    \text{object:\ \ } \mathbf{R}^{-1}\mathbf{K}^{-1}(d\cdot \mathbf{p}^{\mathrm{im}}) & = \mathbf{P}^{\mathrm{world}} = d\cdot(\mathrm{X, Y, Z}) \label{eq:rebuttal-object}\\
    \text{ground:\ \ } \mathbf{R}^{-1}\mathbf{K}^{-1}(\tilde d\cdot \mathbf{\tilde p}^{\mathrm{im}}) & = \mathbf{\tilde P}^{\mathrm{world}} = \tilde d\cdot(\mathrm{\tilde X, \tilde Y, \tilde Z})\label{eq:rebuttal-ground}
\end{align}
}%
where $d$ is the depth value of the point. Here, the point $\mathbf{\tilde p}^{\mathrm{im}}$ in~\cref{eq:rebuttal-ground} is the vertical projection of the ground of $\mathbf{p}^{\mathrm{im}}$ in~\cref{eq:rebuttal-object}. For a given pixel $\mathbf{p}^{\mathrm{im}}$, its corresponding $\mathbf{\tilde p}^{\mathrm{im}}$ can be obtained from our estimated vertical direction (\textit{perspective field}) and the estimated \textit{pixel height}. Note that the world coordinate has its $\mathrm{Z}$ axis pointing vertically upwards, and its $\mathrm{XY}$ plane parallel to the ground plane. The objective is to obtain the location of the reconstructed 3D point $\mathbf{P}^{\mathrm{world}} = d\cdot(\mathrm{X, Y, Z})$, and to eliminate the unknown depth $d$, we need two additional constraints with the help of~\cref{eq:rebuttal-ground}. The \underline{constraint one} is that \textit{all 3D points $\mathbf{\tilde P}^{\mathrm{world}}$ on the ground have a constant $z$-axis value}. Without loss of generality, we assume that the constant is one, to obtain a \textbf{scale-invariant} 3D point cloud. This gives us $\tilde d = 1 / \mathrm{\tilde Z}$, which then leads to the normalized $\mathbf{\tilde P_n}^{\mathrm{world}}$:

{
  \setlength{\abovedisplayskip}{-6pt}
  \setlength{\belowdisplayskip}{6pt}
  \setlength{\abovedisplayshortskip}{0pt}
  \setlength{\belowdisplayshortskip}{3pt}
 \begin{align}
    \mathbf{\tilde P_n}^{\mathrm{world}} = (\mathrm{\tilde X/\tilde Z, \tilde Y/\tilde Z, 1}) = (\mathrm{X_n, Y_n, 1})\label{eq:rebuttal-ground-updated}
\end{align}
}%
Then the \underline{constraint two} is that \textit{the 3D point $\mathbf{P}^{\mathrm{world}}$ and its vertical ground projection $\mathbf{\tilde P}^{\mathrm{world}}$ have identical $\mathrm{XY}$ coordinates}. With this, we know that {\small $d = \frac{\mathrm{X_n}}{\mathrm{X}} = \frac{\mathrm{Y_n}}{\mathrm{Y}}$}. We calculate {\small $d = \frac{\mathrm{X_nY_n}}{\mathrm{XY}}$} for numerical robustness, and the final normalized 3D point is 

{
  \setlength{\abovedisplayskip}{-6pt}
  \setlength{\belowdisplayskip}{6pt}
  \setlength{\abovedisplayshortskip}{0pt}
  \setlength{\belowdisplayshortskip}{3pt}
 \begin{align}
    \mathbf{P}^{\mathrm{world}}_\mathbf{n} = \frac{\mathrm{X_nY_n}}{\mathrm{XY}} \cdot (\mathrm{X, Y, Z})\label{eq:rebuttal-final}
\end{align}
}%
where $\mathrm{X, Y, Z, X_n, Y_n}$ are calculated from~\cref{eq:rebuttal-object,eq:rebuttal-ground,eq:rebuttal-ground-updated}.

\section{Experiments}\label{sec:experiments}

In this section, we conduct extensive qualitative and quantitative experiments to demonstrate the effectiveness and generalizability of \ourwork. We evaluate our model with classic depth estimation metric and point cloud reconstruction metric on both object-centric images and human-centric images.  We show that repurposing two dense representation predictions leads to a very robust 3D reconstruction framework for diverse categories and viewpoints of images.

\begin{figure*}[!t]
    \centering
    \includegraphics[trim=595 485 140 330, clip=True, width=0.98\linewidth]{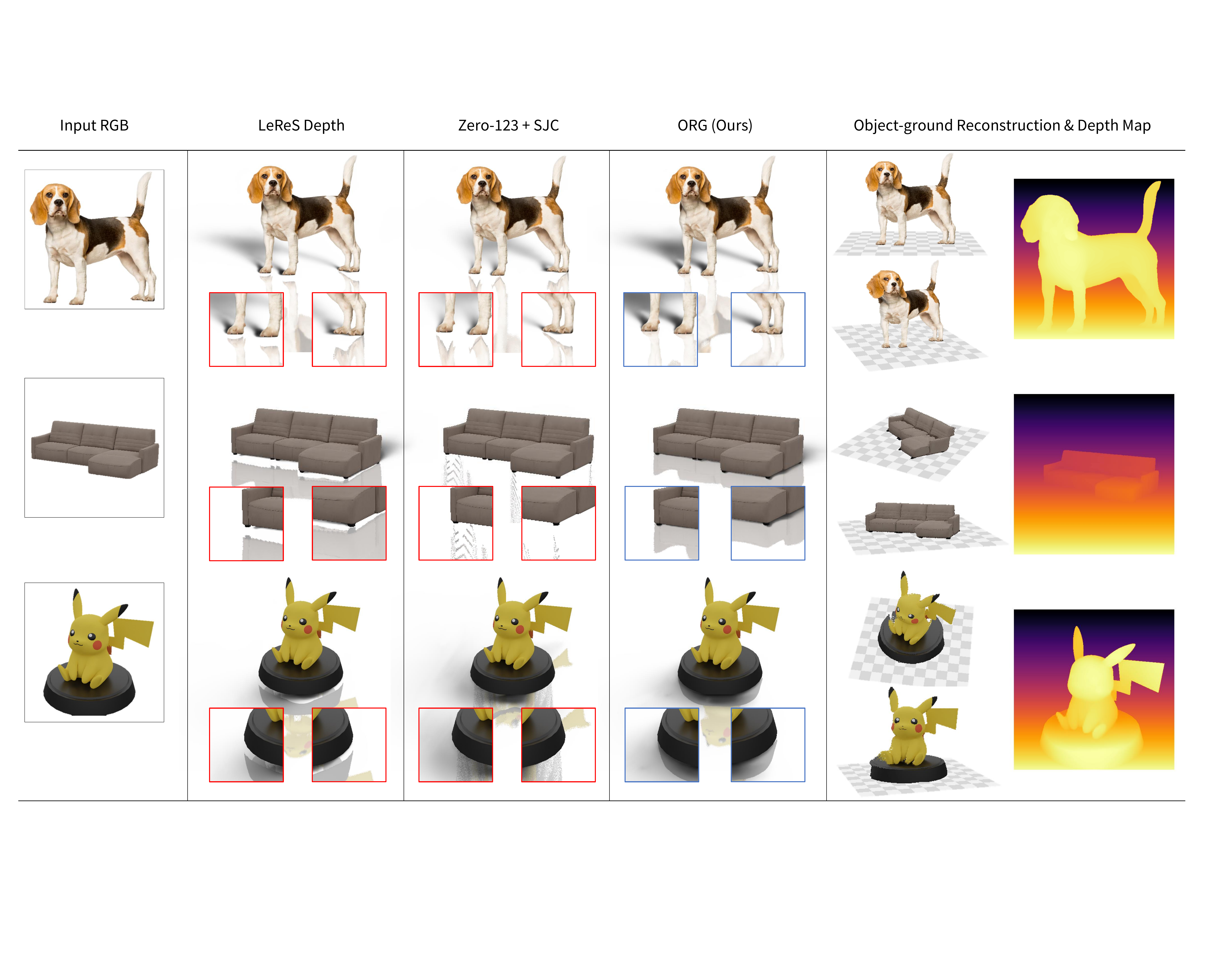}
    \vspace{-2mm}
    \caption{Qualitative results of shadow and reflection generation on the ground, as well as object-ground reconstruction and depth estimation. We show comparison with the depth-based estimation method LeReS~\cite{yin2021leres} and monocular novel view synthesis method Zero-123~\cite{liu2023zero}. \ourwork maintains great object-ground relationship compared with prior methods which leads to much more realistic shadow and reflection generation, as shown in the \textcolor{Myblue}{blue} boxes. Our method runs very fast and can easily output representations like depth map and point cloud.
    }
    \label{fig:final-results}
    \vspace{-3mm}
\end{figure*}

\subsection{Data Rendering} 

Existing object-centric datasets~\cite{reizenstein2021common,ahmadyan2021objectron} do not provide accurate depth map and object-ground rotation information simultaneously. Hence, we render a large-scale dataset from \textit{Objaverse}~\cite{objaverse}. Objaverse is a large-scale object-centric dataset consisting of over 800K high-quality 3D models. For each object in the dataset, we randomly sample 6 sets of camera intrinsic and extrinsic parameters (FoV and rotation matrix), each is used to render an RGB image with pixel height and perspective field ground truth maps. The image dimension is $(512, 512)$. The camera always points at the center of the object and the $z$-axis of world coordinates points orthogonally to the ground plane. We use a physically-based renderer Blender~\cite{blender} to render realistic surface appearance and develop a CUDA-based ray-tracer to efficiently render front and back surface pixel heights. We conduct corrupt data filtering to remove images with incorrect annotations and images with objects that are too small on the canvas. This results in 3,364,052 images in the dataset in total. We split the objects into train/val/test sets in 8:1:1. We also randomize the intensity, position, number of light sources, and distance between camera and object to increase the dataset diversity. We will release our \textit{data rendering script} and \textit{rendered dataset}. More details of the implementation and the dataset are in the supplementary.

\subsection{Baselines}

We compare our method with single-view depth estimation, image-to-3D reconstruction, and camera parameter estimation work. For depth estimation work, we compare with LeReS~\cite{yin2021leres}, MiDaS~\cite{ranftl2020midas,Ranftl2021midasdpt,birkl2023midasv3}, and MegaDepth~\cite{li2018megadepth}, which are single-view generic depth estimation methods pretrained on large-scale datasets. For image-to-3D reconstruction work, we compare with Zero-123~\cite{liu2023zero}, a single-image novel-view synthesis and reconstruction method also pretrained on Objaverse dataset~\cite{objaverse}. For camera parameter estimation, we compare with the state-of-the-art off-the-shelf camera estimator CTRL-C~\cite{lee2021ctrl} and a heuristic method we implemented by eyeballing a rough FoV and pitch angle for all evaluation samples in the test set to get the camera focal length and rotation matrix. Using estimated camera parameters, we can convert the predicted depth map into point clouds. Note that in order to generate depth map and point clouds for objects, we use image mask to remove the background region of our prediction, as well as for existing methods, as we can see in Figure~\ref{fig:final-results}. More details are provided in the supplementary.

\vspace{1.5mm}\mypar{Metrics.} For a fair comparison with existing methods, we evaluate our method on depth estimation and point cloud reconstruction tasks. In the meanwhile, we visualize the estimated ground plane together with reconstruction objects to validate the object-ground correlation. For depth estimation, following previous methods~\cite{ranftl2020midas,yin2021leres}, we use absolute mean relative error (AbsRel) and the percentage of pixels with $\delta_1=\max(\frac{d_i}{d_i^*}, \frac{d_i^*}{d_i})<1.25$. We follow MiDaS~\cite{ranftl2020midas} and LeReS~\cite{yin2021leres} to align the scale and shift before evaluation. For point cloud estimation, following prior work~\cite{chen2020oasis,yin2021leres}, we use Locally Scale Invariant RMSE (LSIV) and Chamfer Distance (CD). In addition, we also evaluate our direct estimation on pixel height, latitude-vector field and up-vector field using mean-square error (MSE) and absolute error (L1).

\setlength{\tabcolsep}{3pt}
\begin{table}[!t]
\normalsize
\centering
\resizebox{0.8\linewidth}{!}{
\begin{tabular}{@{\hskip 1mm}r@{\hskip 4mm}r@{\hskip 4mm}r@{\hskip 4mm}r}
\toprule
& small   & medium     & large               \\ 
\midrule
Baseline        & 0.23                      & 0.37           & 0.72              \\
\ourwork (Ours) & 0.21                      & 0.28           & 0.45              \\
\textit{diff}   & \textcolor{mygreen}{$-0.02$} & \textcolor{mygreen}{$-0.09$}    & \textcolor{mygreen}{$\mathbf{-0.27}$}       \\
\bottomrule
\end{tabular}
}
\vspace{-1mm}
\caption{\ourwork achieves higher improvement against baseline model (DPT-BeiT~\cite{Ranftl2021midasdpt,birkl2023midasv3} + Ctrl-C~\cite{lee2021ctrl}) when objects have larger viewpoint diversity. We report results on point clouds LSIV metrics on validation set. \textit{Small}, \textit{medium}, and \textit{large} stand for different levels of viewpoint diversity of the samples.}
\label{table:ablation-viewpoint}
\vspace{-3mm}
\end{table}

\subsection{Shadow, Reflection, and Reconstruction}

We show results for 3D reconstruction, shadow generation, and reflection generation on unseen objects in~\Cref{fig:final-results}. We compare generation performance with the monocular depth estimation method~\cite{yin2021leres} and the novel view synthesis method~\cite{liu2023zero}. For both methods, we use Ctrl-C~\cite{lee2021ctrl} to predict camera parameters. Since these methods do not model the ground explicitly, we use the estimated pitch angle to obtain the ground plane by assuming that it passes through the object's lowest point (point with smallest height value). For the baseline for novel view synthesis, we use SJC~\cite{wang2023scoresjc} to reconstruct the shape of the object. As depicted in~\Cref{fig:final-results}, notably, there is a marked improvement in the quality of shadows and reflections, particularly at contact points on the ground, as highlighted in the designated boxes. Our research also includes object-ground reconstructions and depth map conversions. The 3D shape of the reconstructed models in our work is not only realistic but also maintains an accurate vertical alignment with the ground plane. This visualization effectively demonstrates our model's versatility, showcasing its exceptional performance across a wide array of object categories, poses, and viewpoints.

\begin{figure}[!t]
    \centering
    \includegraphics[trim=162 90 162 90, clip=True, width=1.\linewidth]{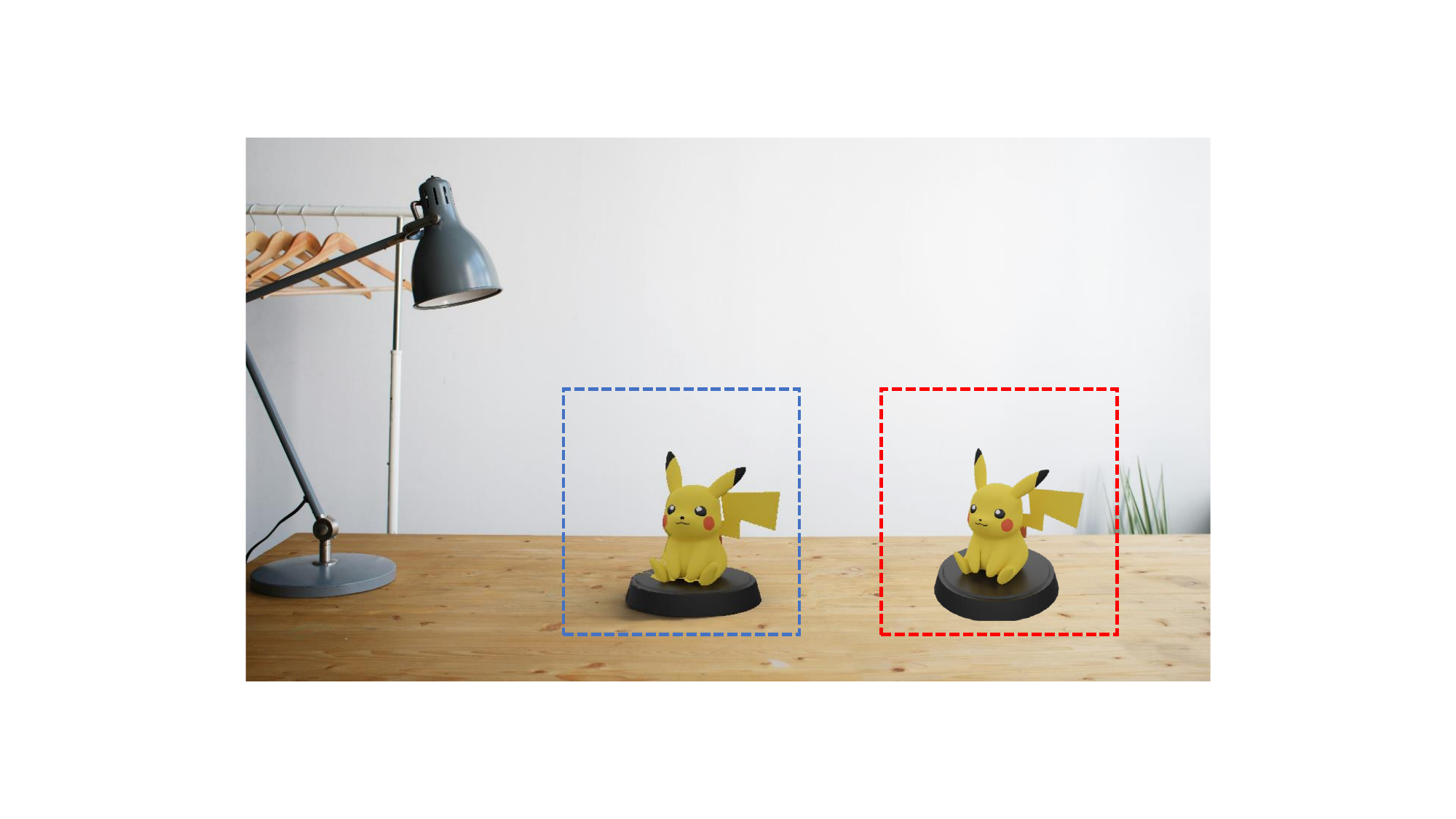}
    \vspace{-6mm}
    \caption{Reconstructed object in the realistic background. \textit{Blue}: novel view synthesis and realistic background composition with our method. \textit{Red}: direct background composition.}
    \label{fig:composition}
    \vspace{-2mm}
\end{figure}

\setlength{\tabcolsep}{3pt}
\begin{table}[!t]
\centering
\resizebox{0.95\linewidth}{!}{
\begin{tabular}{@{\hskip 1mm}l@{\hskip 2mm}r@{\hskip 4mm}r@{\hskip 3mm}cr}
\toprule
& Object Geometry   & Camera Parameters     & \textbf{LSIV}$\downarrow$     & \textit{diff} \\ 
\midrule
& depth                 & OFS estimator                 & 1.25              & $0$  \\
& depth                 & perspective field             & 1.01              & \textcolor{mygreen}{$-0.24$}  \\
& pixel height          & OFS estimator                 & 0.98              & \textcolor{mygreen}{$-0.27$}  \\
& \textbf{pixel height} & \textbf{perspective field}    & \textbf{0.81}     & \textcolor{mygreen}{$\mathbf{-0.44}$}  \\
\bottomrule
\end{tabular}
}
\vspace{-1mm}
\caption{Our proposed joint estimation of pixel height and perspective field contribute to the final performance. We report results on point clouds LSIV metrics. OFS stands for off-the-shelf, and we use Ctrl-C as the OSF estimation in this experiment.}
\label{table:ablation-components}
\vspace{-3mm}
\end{table}

\setlength{\tabcolsep}{3pt}
\begin{table*}[!t]
\centering
\resizebox{0.95\linewidth}{!}{
\begin{tabular}{@{\hskip 1mm}l@{\hskip 2mm}c@{\hskip 4mm}cc@{\hskip 3mm}cc@{\hskip 5mm}c@{\hskip 2mm}c@{\hskip 2mm}c@{\hskip 1mm}}
\toprule
&  & \multicolumn{2}{c}{Depth Map\ \ \ \ \ } & \multicolumn{2}{c}{Point Clouds\ \ \ \ \ \ \ \ \ \ } & {Pixel Height\ } & {Lati-Vector\ } & {Up-Vector\ \ } \\ 
\cmidrule(r{2.5mm}){3-4}
\cmidrule(l{0mm}r{6.5mm}){5-6}
\cmidrule(l{0mm}r{2mm}){7-7}
\cmidrule(l{0mm}r{2mm}){8-8}
\cmidrule(l{0mm}r{2mm}){9-9}
&  camera parameters  & \textbf{AbsRel}$\downarrow$ & $\boldsymbol{\delta_1}$$\uparrow$ & \textbf{LSIV}$\downarrow$ & \textbf{CD}$\downarrow$ & \textbf{L1}$\downarrow$ & \textbf{L1}$\downarrow$ & \textbf{L1}$\downarrow$ \\
\midrule
{MegaDepth~\cite{li2018megadepth}}& \multirow{5}{*}{heuristic const.}
    & 39.4  & 53.7  & 1.60  & 1.73  & 36.8  & \multirow{5}{*}{8.77}     & \multirow{5}{*}{3.02}  \\
{NDDepth~\cite{shao2023nddepth}}      & 
    & 35.8  & 54.2  & 1.49  & 1.65  & 30.9  &     &     \\
{MiDaS~\cite{ranftl2020midas,Ranftl2021midasdpt,birkl2023midasv3}}      & 
    & 22.7  & 77.9  & 1.31  & 1.45  & 26.0  &     &     \\
{LeReS~\cite{yin2021leres}}         &      
    & 30.0  & 63.1  & 1.11  & 1.34  & 24.5   &     &     \\
{\ourwork (Ours)}      & 
    & 24.6  & 71.2  & 1.07  & 1.39  & 15.4     &     &     \\
\midrule
{MegaDepth~\cite{li2018megadepth}}& \multirow{5}{*}{Ctrl-C~\cite{lee2021ctrl}}
    & 39.4  & 53.7  & 1.51  & 1.64  & 31.1   & \multirow{5}{*}{5.45}     & \multirow{5}{*}{1.79} \\
{NDDepth~\cite{shao2023nddepth}}      & 
    & 35.8  & 54.2  & 1.46  & 1.60  & 28.3  &     &     \\
{MiDaS~\cite{ranftl2020midas,Ranftl2021midasdpt,birkl2023midasv3}}      & 
    & 22.7  & 77.9  & 1.22  & 1.39  & 20.7     &     &     \\
{LeReS~\cite{yin2021leres}}         &      
    & 30.0  & 63.1  & 1.05  & 1.31  & 20.4   &     &     \\
{\ourwork (Ours)}                   &            
    & 21.1  & 76.0  & 0.99  & 1.27     & 15.4   &     &     \\
\midrule
{\textbf{\ourwork} (Ours)}   & \textbf{Ours} 
    & \textbf{19.1}  & \textbf{81.2}  & \textbf{0.93} & \textbf{1.26}  &  \textbf{15.4} & \textbf{4.94}    & \textbf{1.45}  \\
\bottomrule
\end{tabular}
}
\vspace{-1mm}
\caption{\ourwork perform consistently the best in both depth estimation and point cloud estimation tasks of object-centric images under all metrics. We use two off-the-shelf camera estimation algorithms to make up for the unknown camera parameters. Pixel Height metric is reported in absolute error of number of pixels, Latitude-vector Field and Up-Vector Field are reported in degrees.}
\label{table:obj-centric-dp-pc}
\vspace{-2mm}
\end{table*}

\subsection{Novel View Synthesis and Image Composition}

We demonstrate applications such as object view manipulation, shadow generation, and image composition in~\Cref{fig:composition}. In the red box, we show direct copy-and-paste composition as a comparison, and performance of \ourwork in shown in the blue box. We notice that the simple copy-pasting method does not match the camera perspective of the new object and its supporting plane in the background, creating unrealistic visual effects. Our method, on the other hand, estimate the background perspective, reconstruct the object into 3D and re-render it from the target perspective, and generate photo-realistic shadow from the estimated object shape, achieving better visual alignment and realism.

\begin{figure}[!t]
    \centering
    \includegraphics[trim=1005 835 1150 555, clip=True, width=\linewidth]{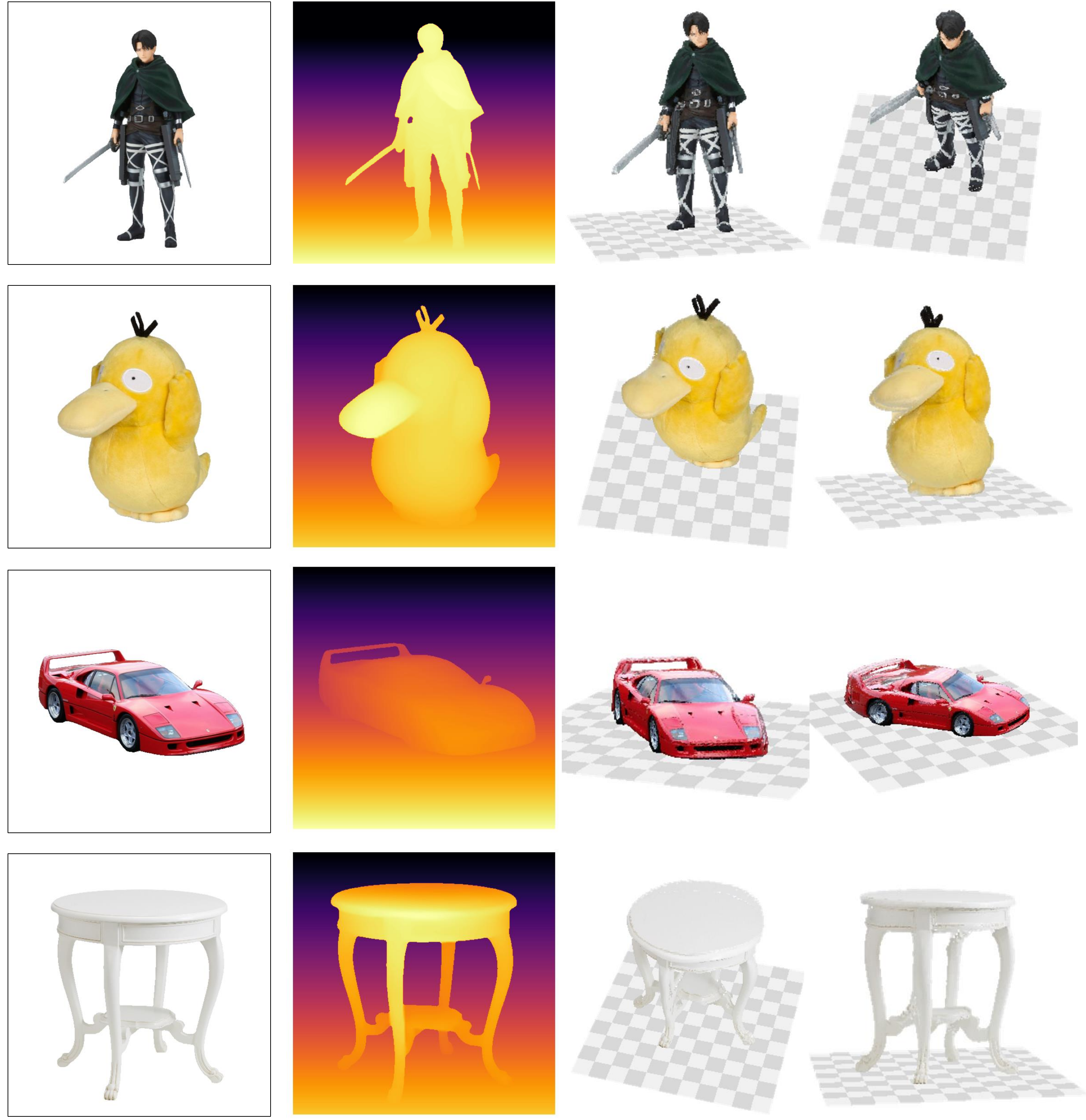}
    \vspace{-5mm}
    \caption{More qualitative results of \ourwork in depth map generation and object-ground reconstruction. Our method generalizes well to unseen in-the-wild images.
    }
    \label{fig:more-ours}
    \vspace{-3mm}
\end{figure}

\vspace{1.5mm}\mypar{More Qualitative Results.} Moreover, ~\Cref{fig:more-ours} illustrates additional qualitative results from our study, focusing on depth map generation and object-ground reconstruction. Our methodology exhibits remarkable proficiency in reconstructing ground-supported objects across various types, underscoring the robustness of our approach.

\subsection{Object with Diverse Viewpoints}

We also break down the evaluation into subsets of samples with different range of camera angles. More specifically, we divide the difficulty level by the pitch angle because natural images usually have more diverse pitch angles but close to zero roll angles. Taking the mean pitch angle of the entire dataset, samples with a pitch angle smaller than 10 degrees difference than the mean angle are marked as \textit{small} viewpoint diversity. Samples with a pitch angle difference between 10 and 30 degrees are marked as \textit{medium} viewpoint diversity, and samples with pitch angle difference greater than 30 degrees are marked as \textit{large} viewpoint diversity. Results in Table~\ref{table:ablation-viewpoint} show that \ourwork achieves a higher improvement compared to the baseline model (LeReS~\cite{yin2021leres} + Ctrl-C~\cite{lee2021ctrl}) when objects have greater viewpoint diversity. This is because the traditional viewpoint estimation model struggles for object-centric images, especially for samples with extreme pitch angles.

\subsection{Importance of Joint Estimation}

The results in Table~\ref{table:ablation-components} show that the joint learning of pixel height and perspective field leads to the best reconstruction performance compared to the depth estimation and off-the-shelf camera parameter estimator. More specifically, without modifying the model architecture, we change the objective of our model from pixel height estimation to depth estimation following the loss used in LeReS~\cite{yin2021leres}. Trained with the same dataset and scheduler, the pixel height representation is able to achieve better point-cloud reconstruction than depth-based learning. We argue that this is because the representation focuses more on object-ground geometry rather than object-camera geometry, which is more natural and easier to infer from object-centric images. This observation further validates that the superior generalizability of \ourwork comes from the better representation design and the joint training strategy, rather than the dataset.

\subsection{Qualitative Evaluation on Reconstruction}

We compare the depth map estimation, point cloud generation, and the prediction of our representations with four state-of-the-art monocular depth estimation and 3D reconstruction methods on the held-out test set. We use the state-of-the-art camera parameter estimation model Ctrl-C~\cite{lee2021ctrl} and a heuristic estimation to compensate for missing intrinsic and extrinsic information from previous methods. We convert the raw output into depth map and point clouds for a fair comparison with existing methods.  
As shown in Table~\ref{table:obj-centric-dp-pc}, our method performs consistently the best in both depth estimation and point cloud estimation tasks for object-centric images under all metrics. We also try using the other two alternative camera parameter estimators to reconstruct the point cloud from the pixel height estimation. And we can see that using the same off-the-shelf camera estimator, \ourwork can still outperform existing methods on both two tasks. We make sure that no samples in the evaluation dataset are seen by prior methods or our method during the training phase, in order to create a zero-shot evaluation scenario. Results show that \ourwork achieves a great generalization ability in the object-centric 3D reconstruction task.

Furthermore, we also break down the evaluation into pixel height, latitude vector, and up-vector estimation, and evaluate with mean absolution error in the generic space of all three predictions (number of pixels for pixel height and degrees for two perspective fields). For prior methods, we use Ctrl-C and the heuristic constant (by grid search) to estimate elevation angle, roll angle and camera FoV, and convert them into perspective field representations for comparison. Their pixel height estimations are also converted using depth estimation and camera parameter estimations. As we can see in Table~\ref{table:obj-centric-dp-pc}, our method outperforms the baselines in all three tasks. These experiments demonstrate the robustness and generalizability of \ourwork over prior methods in object 3D estimation and reconstruction.

\section{Conclusion}\label{sec:conclusion}

In this paper, we proposed \ourwork, to our knowledge, the first data-driven architecture that simultaneously reconstructs 3D object, estimates camera parameter, and models the object-ground relationship from a monocular image. To achieve this, we propose a new formulation for representing objects in relation to the ground. Qualitative and quantitative results on unseen object and human datasets as well as web images demonstrate the robustness and flexibility of our model, which marks a significant step towards in-the-wild single-image object geometry estimation.

{
    \small
    \bibliographystyle{ieeenat_fullname}
    \bibliography{main}
}

\clearpage
\setcounter{page}{1}
\maketitlesupplementary

\renewcommand{\thesection}{\Alph{section}}
\renewcommand{\thefigure}{\Alph{figure}}
\setcounter{section}{0}
\setcounter{figure}{0}

\section{Implementation Details} \label{sec:supp-dataset}

Here we provide more details regarding the implementation and training of our model. 

\vspace{1.5mm}
\mypar{Backbone and Decoder.} We use PVTv2-b3~\cite{wang2021pyramid} pretrained on COCO dataset~\cite{lin2014microsoft} as our encoder backbone. And we use a decoder of a similar design as SegFormer~\cite{xie2021segformer}, which consists of four multi-layer-perceptron (MLP) layers to extract feature maps of different scales. We predict two dense fields with five channels: two for front and back surface pixel height map, one for latitude field, and two for gravity field.

\vspace{1.5mm}
\mypar{Data Normalization.} For pixel height estimation, we normalize the ground truth maps by dividing them with the height of the image, which roughly turns the range of the pixel heights into $[0, 1]$ such that our model is not affected by objects at different scale. For two perspective fields, we normalize the latitude field into $[0, 1]$ and we represent the gravity (up-vector) field with a (sine, cosine) tuple as described in~\Cref{sec:approach:net} in the main paper. The estimation of all three representations are formulated as regression problems and trained by MSE loss. Similar to existing methods~\cite{ranftl2020midas,yin2021leres,saito2019pifu,saito2020pifuhd}, due to the estimation of a normalized pixel height representation, our reconstructed models (~\Cref{sec:approach:geometry}) preserve the 3D geometry of the original objects but are scale-ambiguous. We calibrate the objects reconstructed by our methods and prior method using a linear scaling following LeReS~\cite{yin2021leres}. 

\vspace{1.5mm}
\mypar{Objact Mask.} All the datasets we utilized for training and quantitative evaluation come with object masks, which are from either human annotation or off-the-shelf segmentation models. When evaluating on web images, we utilize segmentation model Rembg with u2net backbone~\cite{rembg} to get the foreground mask.

\vspace{1.5mm}
\mypar{Data Generation.} We use the physically-based rendering engine Blender~\cite{blender} to render realistic RGB channel results.
The front and back surface pixel height is calculated by our ray tracer.  
In detail, we shoot one ray to each pixel, find the first and last ray-object intersection points and calculate their relevant 3D foot points ($z$=0).   
Then we project the intersection points and their footpoints onto the camera. 
The pixel heights are calculated by measuring the distances of the projected intersection points and their projected foot points in pixel units.
Our pixel height calculation is efficient and can be computed in real time.

\vspace{1.5mm}
\mypar{Training and Scheduling.} The model is trained with AdamW~\cite{loshchilov2017decoupled} optimizer with initial learning rate $0.0005$ and weight decay 1$e$-2 for $60$K steps with batch size 8 on a 4-A100 machine. We schedule the multi-step training stages at step $30$K, $40$K, and $50$K, with learning rate decreases 10$\times$ each time. We resize the images to $(512, 512)$ resolution. We use horizontal flipping, random cropping, and color jittering augmentation during training. And because horizontal flipping, random cropping and resizing will affect the values of our representations, we update the ground truth maps accordingly. The whole model is implemented using the PyTorch framework~\cite{paszke2019pytorch}.

\section{More Qualitative Analysis}

Here we demonstrate more visualization examples of \ourwork. We show more diverse categories of objects with different camera viewpoints on random web images, and also full object geometry reconstruction results.

\begin{figure*}[!t]
    \centering
    \includegraphics[trim=620 315 620 200, clip=True, width=\linewidth]{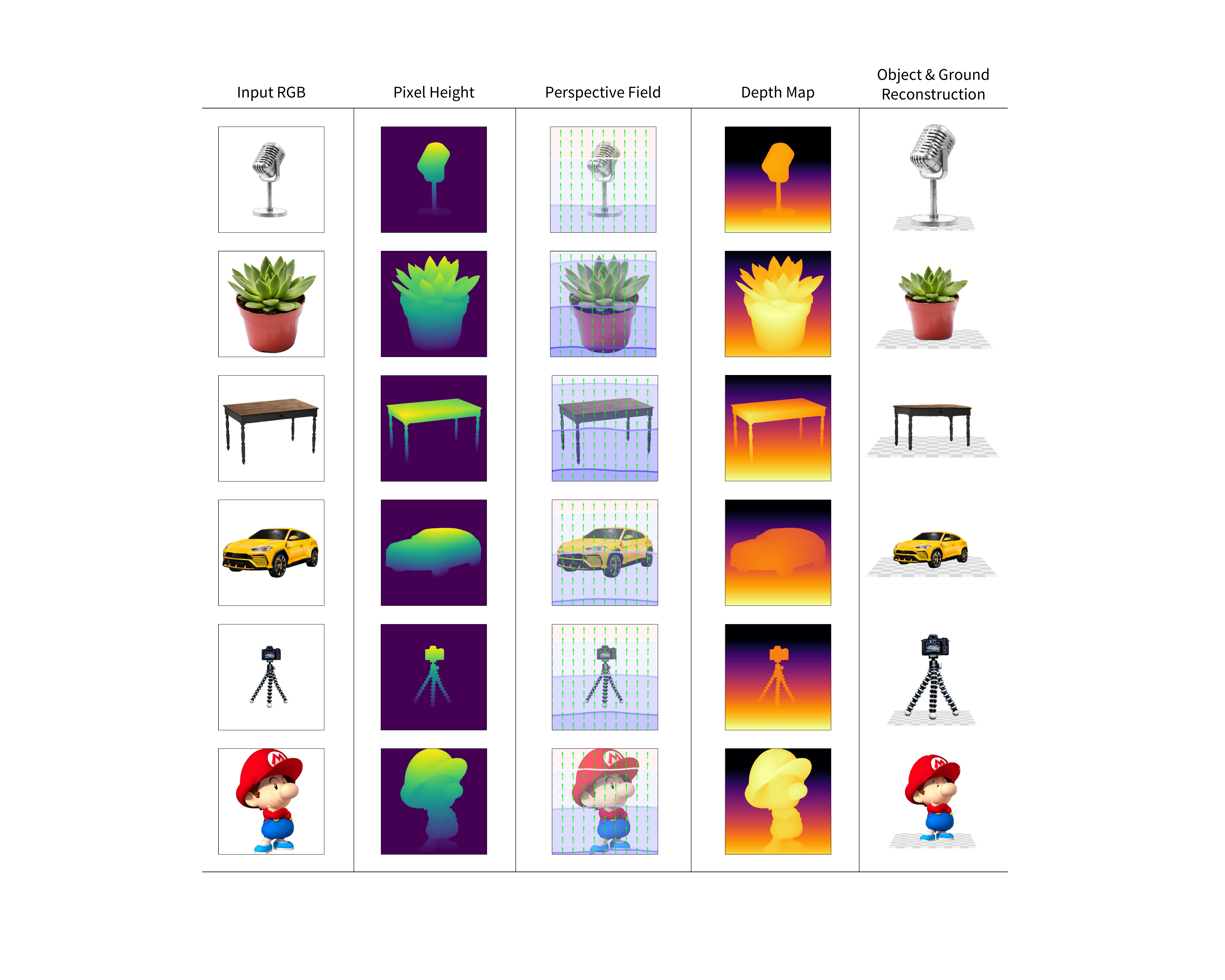}
    \vspace{-5mm}
    \caption{Visualization of \ourwork on pixel height, (foreground) perspective fields, depth map, and object-ground reconstruction results. The results demonstrate that our work generalizes to various categories of objects.
    }
    \label{fig:supp-webobj}
    \vspace{-3mm}
\end{figure*}

\vspace{1.5mm}
\mypar{Diverse Categories.} In~\Cref{fig:supp-webobj}, we show our direct estimation of pixel height and prospective fields, and also visualize the reprojected depth maps and reconstructed object-ground point clouds of diverse categories of objects from web images. The categories include common objects like microphone, plant, car, and tripod, as well as cartoon figures. The results show great generalizability and robustness of our method in the wild.

\begin{figure*}[!t]
    \centering
    \includegraphics[trim=325 950 370 950, clip=True, width=\linewidth]{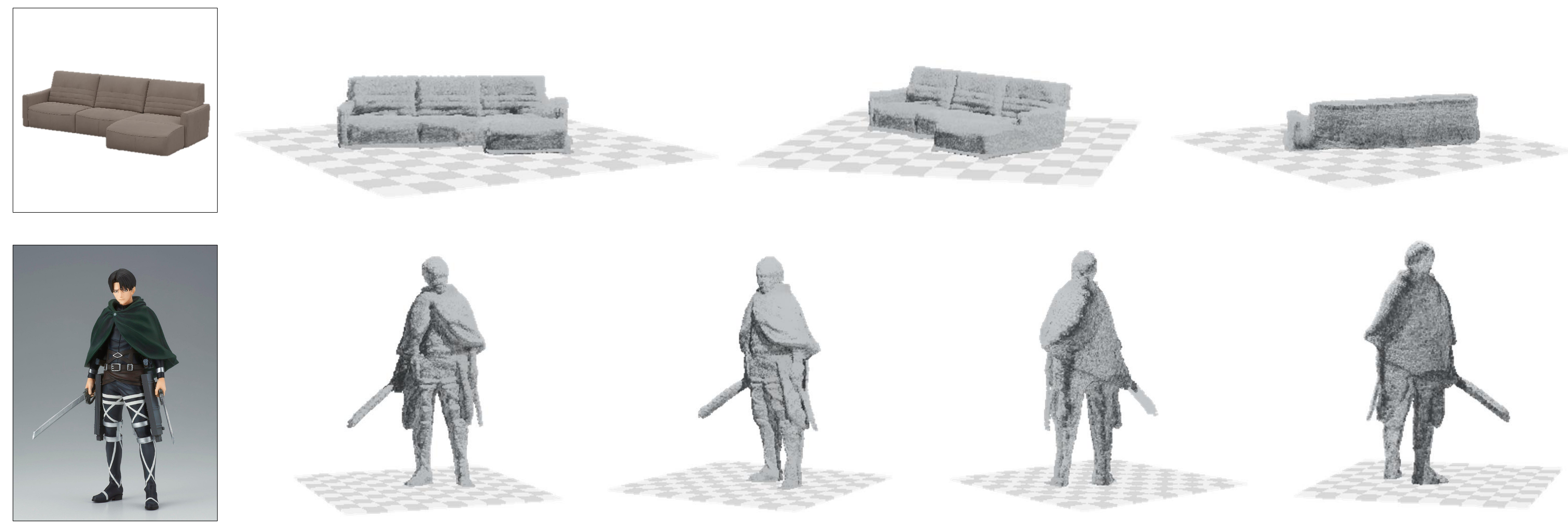}
    \vspace{-5mm}
    \caption{Visualization of \ourwork on the full object geometry (front surface and back surface) with the ground plane.
    }
    \label{fig:supp-full3drecon}
    \vspace{-3mm}
\end{figure*}

\vspace{1.5mm}
\mypar{Object-Ground Reconstruction.} In addition to our previous analyses, we present a detailed visualization of the complete 3D geometry of reconstructed objects and ground in~\Cref{fig:supp-full3drecon}. Here, the objects are represented using 3D point clouds. Despite employing a simplified geometric model in our approach, our results effectively showcase superior reconstruction quality, particularly for objects with relatively straightforward geometric structures. This aspect of \ourwork highlights the balance between model simplicity and the ability to achieve high-fidelity reconstructions, even with less complex geometries.

\section{Limitations and Future Work}

Primarily, our approach relies on a simplified object shape assumption, optimizing for efficient image editing (\eg, reflection, shadow generation, and ground-aware object pose change). However, this simplification may yield less than satisfactory 3D reconstruction results for objects with intricate geometries, particularly in estimating their back surfaces. Additionally, our method focuses solely on the geometric aspects of objects, excluding considerations of color and texture. We propose that leveraging our estimated geometry as a conditioned prior could significantly enhance image-inpainting processes, presenting a promising direction for future research.

\end{document}